\documentclass[preprint,12pt]{elsarticle}

\usepackage{graphicx}
\usepackage{amssymb}

\usepackage{url}

\usepackage{breakurl}
\usepackage[breaklinks]{hyperref}

\usepackage{float}

\usepackage{acronym} 

\acrodef{PWM}[PWM]{Pulse Width Modulation}
\acrodef{PFM}[PFM]{Pulse Frequency Modulation}
\acrodef{FPGA}[FPGA]{Field Programmable Gate Array}
\acrodef{PID}[PID]{Proportional Integral and Derivative}
\acrodef{DVS}[DVS]{Dynamic Vision Sensor}
\acrodef{VLSI}[VLSI] {Very Large Scale Integration}
\acrodef{AER}[AER]{Address Event Representation}
\acrodef{PCB}[PCB]{Printed Circuit Board}
\acrodef{FSM}[FSM]{Finite State Machine}
\acrodef{LUT}[LUT]{Look-Up Table}
\acrodef{CPG}[CPG] {Central Pattern Generator}
\acrodef{sCPG}[sCPG] {Spiking Central Pattern Generator}
\acrodef{SNN}[SNN]{Spiking Neural Network}
\acrodef{ANN}[ANN]{Artificial Neural Network}
\acrodef{NN}[NN]{Neural Network}
\acrodef{MLP}[MLP]{Multi Layer Perceptron}
\acrodef{DOF}[DOF]{degrees of freedom}
\acrodef{HDL}[HDL]{Hardware Description Language}
\acrodef{VHDL}[VHDL]{VHSIC Hardware Description Language}
\acrodef{FW}[FW]{Forward}
\acrodef{BW}[BW]{Backward}
\acrodef{FRS}[FRS]{Force Resistive Sensor}
\acrodef{LIF}[LIF]{Leaky Integrate and Fire}
\acrodef{HBP}[HBP]{Human Brain Project}
\acrodef{ML}[ML]{Machine Learning}
\acrodef{LSTM}[LSTM]{Long-Short Term Memory} 
\acrodef{RNN}[RNN]{Recursive Neural Networks}
\acrodef{AI}[AI]{Artificial Intelligence}
\acrodef{CNN}[CNN]{Convolutional Neural Networks}
\acrodef{XAI}[XAI]{Explainable Artificial Intelligence} 

\usepackage{ulem}
\usepackage{xcolor}

\journal{Expert Systems with Applications}

\begin{document}

\begin{frontmatter}


\title{Real-time detection of uncalibrated sensors using Neural Networks}



\author[1,3]{Luis J. Muñoz-Molina}
\author[1]{Ignacio Cazorla-Piñar}
\author[2]{Juan~P.~Dominguez-Morales}
\author[3]{Fernando Perez-Peña}

\address[1]{Altran Innovation Center for Advance Manufacturing}
\address[2]{Robotics and Technology of Computers Lab., Universidad de Sevilla, Seville, 41012, Spain.}
\address[3]{School of Engineering, Universidad de Cadiz, Spain.}

\begin{abstract}
Nowadays, sensors play a major role in several contexts like science, industry and daily life which benefit of their use.   
However, the retrieved information must be reliable. Anomalies in the behavior of sensors can give rise to critical consequences such as ruining a scientific project or jeopardizing the quality of the production in industrial production lines. One of the more subtle kind of anomalies are uncalibrations. An uncalibration is said to take place when the sensor is not adjusted or standardized by calibration according to a ground truth value. 
In this work, an online machine-learning based uncalibration detector for temperature, humidity and pressure sensors was developed.
This solution integrates an Artificial Neural Network as main component which learns from the behavior of the sensors under calibrated conditions. Then, after trained and deployed, it detects uncalibrations once they take place.
The obtained results show that the proposed solution is able to detect uncalibrations for deviation values of 0.25º, 1\%~RH and 1.5~Pa, respectively. This solution can be adapted to different contexts by means of transfer learning, whose application allows for the addition of new sensors, the deployment into new environments and the retraining of the model with minimum amounts of data.
\end{abstract}


\end{frontmatter}

\section{Introduction}
\label{S:Intro}

Nowadays, sensors play a major role in the connected world \cite{FutureIoTNetworks, MEMS, RolesChallengesNetworkSensors, iotpharma}. Science, industry or even day-to-day devices integrate sensors to collect the information coming from the surrounding environment. Ensuring the reliability and consistency of the information collected is essential in order to guarantee the proper use of the information.  
The reliability of the acquired information depends on the capabilities of the sensor, but also on its calibration status. An uncalibration is said to take place when the sensor is not adjusted or standardized by calibration according to a ground truth value. In general, anomalies can appear in several shapes \cite{faultstypes}, however the uncalibration event usually appear in the form of long-term drifts with different kind of responses such as linear, exponential, logarithmic or, simply, an irregular drift. Furthermore, there are cases where the magnitude to control is not directly controlled, that is, the set point is unknown. Thus, in the latter case, the detection of potential uncalibrations can be really challenging. As it happens with some other kind of anomalies in sensor behaviour \cite{tobar2011anomaly}, the detection of uncalibrations is a key issue in order to ensure fundamental aspects such as reliability, safety or cost-effectiveness in crucial assets in different contexts, for instance, pharma, energy, aeronautics, etc. All these properties, make the proposed solution a great candidate to be deployed in a wide variety of context of both scientific and industrial and, more precisely, in the context of industry 4.0.


The uncalibration detection problem can be tackled from different perspectives or paradigms. The available information is a key factor in order to determine what perspective should be used. On the one hand, the approaches based on the use of classical techniques can be considered. This is the case of similarity-based modeling and multivariate analysis \cite{lamrini2018anomaly} or the time series analysis \cite{feremans2019pattern}. A survey of some other classical approaches for anomaly derection can be seen in \cite{surveyanomalydetectMLandStatis}. It is important to note that the techniques mentioned in the previous works were used for the detection of specific kinds of anomalies, not uncalibrations. 

On the other hand, due to the huge development of \ac{ML} and \ac{AI} experienced during the last years, the use of techniques and approaches based on them has increased \cite{SurveyDLAnomaly, surveyanomalydetectMLandStatis}. An example of this fact is the case of \cite{ahmad2017unsupervised}, where anomalies, defined as an unusually different behaviour at a specific time, were detected by means of the use of Hierarchical Temporal Memory. Another interesting example can be seen in \cite{liu2019sensor}, where a solution based on \ac{CNN} was developed in order to detect sensor failures. An example of the application of Neural networks for anomaly detection in power plants can be found in \cite{MLinPWERPLANTS}. Finally,an overview of Neural Network applications to fault diagnosis and detection in engineering-related systems can be seen in \cite{NeuralNEtworkFaultDiagnosis}. As in the classical approaches, it is important to note that all those developments were devoted to specific anomalies detection, not uncalibration detection in the sensor system.

In this work, an online uncalibration detector for temperature, humidity and pressure sensors under unknown conditions (set points unknown) has been developed. The major novelty of the paper is the use of an \ac{ANN} as an expert on the sensors behaviour, so that the estimated behaviour by the \ac{ANN} can be compared to the one measured and thus, potential uncalibrations can be detected. Different statistical techniques, such as goodness of fit and confidence intervals, are used in order to check that the estimation of the neural network can be considered as equal or different from the obtained measures.

The developed architecture detects the ongoing uncalibrations and it exhibits a great flexibility and scalability. Indeed, the obtained results suggest that the solution can be trained with a generic set of sensors and then be specialized for more specific contexts with a much lower amount of data. Thus, for instance, the solution could be trained with a standard set of sensors and then be specialized for a more specific set devoted to a more concrete task. Also, including new sensors into the solution is very easy, and fairly low amount of data is required. Finally, retraining the solution requires data coming from about one week of data (at a sampling rate of one sample per minute and per sensor).

The proposed detector has been tested in a real context in the facilities of a company of the pharmaceutical sector. Thus, it is important to note that the presented work has been developed under the production conditions of the aforementioned company.


The rest of the paper is structured as follows: section~\ref{S:Materials} introduces the materials used in this work, including the dataset~(\ref{subsec:dataset}). The implemented methods are described in section~\ref{sec:methods}, together with the \ac{NN} model. Then, the results obtained are presented in section~\ref{sec:results}, which are divided in two different experiments: first, the performance of a proposed \ac{NN} model is evaluated (\ref{subsec:mlexper}), and then its scalability and flexibility is evaluated on (\ref{subsec:tlexper}). Section~\ref{sec:discussion} presents the conclusions and the discussion of this work, respectively.

\section{Materials}
\label{S:Materials}

This section describes the sensors used and the dataset obtained from them.

\subsection{Sensors}
\label{subsec:sensors}
Three different variables were measured, namely, temperature, humidity and pressure. Temperature and humidity sensors are integrated within a single device, the Novasina nSens-HT-ENS. The humidity measurement range of this device goes from $0\%$RH to $100\%$ RH, with a measurement accuracy of $0.5\%$ RH. Regarding the temperature measurement range, it goes from -20\textdegree to +80\textdegree, with a measurement accuracy of 0.1\textdegree. The sensor used to measure the pressure is the Novasina Pascal-ST-ZB. It is a differential pressure sensor based on static pressure measurement. Regarding its measurement range, it goes from -50 Pa to 50 Pa, with a measurement accuracy of $<0.5\%$ fs (full scale).

    
    

    

\subsection{Dataset}
\label{subsec:dataset}
The dataset used to obtain the results presented in this work consists of temperature, humidity and pressure values coming from the sensors installed across different clean rooms devoted to drug production. Rooms are located in two different floors of a building: the second floor and the basement. More precisely, the data coming from floor 2 is described in Table \ref{trainingdataset}. This dataset contains data collected during a whole year in different rooms of the second floor. Each room is endowed with only one sensor per category (one temperature/humidity sensor and one pressure sensor). Thus, sensor redundancy is not available. The sensors used are described in \ref{subsec:sensors}. The data was collected at a sample per minute rate during the aforementioned period of time. 

A smaller dataset containing data from rooms located in the basement was also collected. This data is shown in table \ref{trainingdataset}. Conversely to the second floor, the basement dataset is smaller due to the difficulties to access the data generated within this floor. This dataset contains, approximately, one hundred samples per sensor, coming from a total of eleven sensors per category. Besides, samples are not equally distributed. The data points (also called samples) were collected during different periods of the year. 

\begin{table}[!ht]
\caption{Dataset summary. Note that -1 floor refers to the basement.}
\centering
\begin{tabular}{|c||c||c||c|}
\hline
Floor & Sensor Type & Number of sensors & Number of samples\\
\hline
2 & Temperature & 17 & 8935200\\
\hline
2 & Humidity & 17 &8935200\\
\hline
2 & Pressure & 24 & 12614400\\
\hline
-1 & Temperature & 11 & 1100000\\
\hline
-1 & Humidity & 11 & 1100000\\
\hline
-1 & Pressure & 11 & 1100000\\
\hline
\end{tabular}
\label{trainingdataset}
\end{table}

Since the dataset obtained from the second floor is more complete, it was used as the main dataset. Thus, experiments in subsection \ref{subsec:mlexper} and \ref{subsec:tlexper} were conducted taking this dataset for training purposes. However, in subsection \ref{subsec:tlexper}, the dataset coming from the basement was used for testing the adaptation of the solution proposed to a different environment.

Regarding the dataset slicing, two approaches were defined depending on the kind of experiments that were carried out. On the one hand, in subsection \ref{subsec:mlexper}, 60\% of the samples of the data coming from the second floor were used as training set for each type of sensor, 10\% were used as validation set and the remaining 30\% were used as testing set. On the other hand, in subsection \ref{subsec:tlexper}, 70\% of the samples were used for training the main Neural Network, 10\% as testing set and 20\% for the application of the Transfer Learning. In this last case not only data coming from the second floor was used, but also data coming from the basement was also included in order to test the transfer learning application from one floor to another.

The use of the different slices is more detailed in \ref{sec:methods}.

\section{Methodology}
\label{sec:methods}
In this section, the architecture of the implemented solution, which is capable for detecting sensor uncalibration, is presented.

\begin{figure}[h!]
\centering\includegraphics[width=\linewidth, keepaspectratio]{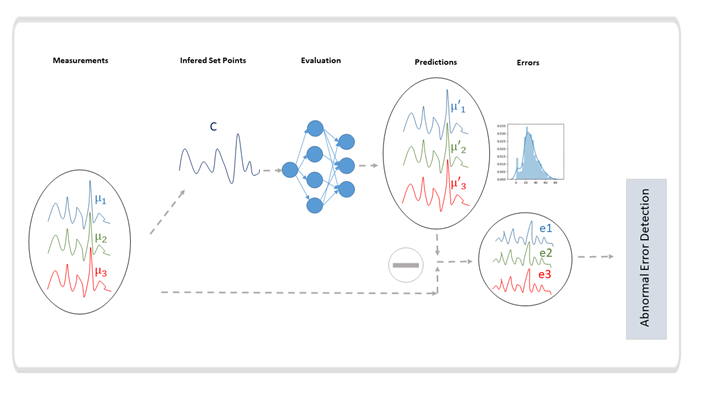}
\caption{Block diagram of the approach presented. Five main stages can be observed: the measurement stage, the estimation of the set point, the evaluation of the estimated set point by the \ac{NN} and the generation of the prediction and, finally, the computation of the error with respect to the real value.}
\label{fig:general_diagram}
\end{figure}

The approach is based on \ac{ML} techniques and it follows the block diagram shown in Figure \ref{fig:general_diagram}. The architecture is based on an \ac{ANN} which, during the training phase, learns how the sensors behave. Then, in the inference step, the \ac{ANN} is able to predict what the measurements should be based on the input stimuli. Eventually, these predictions are compared with the real measurements. If there exists any discrepancy between the prediction and the real measurements, an uncalibration is taking place.

\subsection{Preprocessing}
\label{subsec:preprocessing}

The data used to both train and test the network have to be pre-processed. This preprocessing, using filtering techniques, is a standard procedure to detect anomalous values. These values could yield undesired effects on the performance of the algorithm if they are not properly detected and treated. Thus, this preprocessing phase can be divided into three stages, namely, a threshold-based filtering, the Mahalanobis distance \cite{mahalanobis1936generalized} and the Savitzky-Golay filter \cite{savitzky1964smoothing}.

The threshold-based filtering is based on the thresholds defined for the warning and alarm systems in the rooms. This alarm system is designed to detect extreme values in the incoming measurements. Thus, potential sudden failures in the sensor systems can be detected. It has been observed that these extreme values add noise to the input signal and, therefore, worsen the \ac{NN} learning. These thresholds were used during this research as a way to discard all of those extreme values that were already detected by the system and which could have a deep impact in the \ac{NN} learning process. Thus, a measured value is dropped whenever one of the aforementioned thresholds is exceeded. In this way, only calibrated and non-extreme uncalibrated values remain in the signal.

Next, a filter based on the Mahalanobis distance \cite{mahalanobis1, mahalanobis1936generalized, mahalanobis3} is applied. Mahalanobis distance is an outlier detection method specifically designed for dealing with multidimensional distributions. More precisely, this stage focuses on determining whether an observation including information from the whole set of sensors (i.e a vector with as many entries as sensors) can be considered as an outlier. Thus, in this stage, non-extreme values whose relative occurrence is low would be classified as outliers. 

Finally, the third stage is focused on reducing the noise in the incoming signal. Different kind of filters such as Butterworth or wavelet filters could have been applied in this stage. However, the incoming signal presents a wide variety of different behaviors along the data collection period. Thus, a more flexible noise filtering process is required. The Savitzky-Golay's filter \cite{savitzky1964smoothing}, which computes local polynomial regressions providing new values for each point, is used. This filtering process preserves essential properties of the signal, such as maximum and minimum values or trends, while noise is efficiently dropped out.

After the data obtained from the sensors is preprocessed, the average value of the measurements, at a time, of all the sensors involved is used as input to a \ac{ANN}.

\subsection{Neural network}
\label{subsec:neural_network}

The proposed \ac{ANN} model is a \ac{MLP} that consists of an input layer with one neuron, five hidden layers with three hundred neurons each and an output layer with as many output neurons as sensors in used. After training the network with the average value of the preprocessed data from the sensors, the network is able to decompose the input stimulus into the predicted value for each sensor, which is the expected behavior that each sensor should have based on that input stimuli.  

\subsection{Error computation} 
This is the final step of the proposed approach. In this stage, the difference between the real measured value obtained from the sensors and the ones predicted is computed, evaluated and classified as normal or anomalous. This evaluation is based on the work of \cite{ahmad2017unsupervised}, where Hierarchical Temporal Memory \cite{hawkins2004intelligence} was used to predict the next measurement. That prediction was then compared to the real measurement and residuals were computed. 

In our approach, goodness of fit tests were applied on the errors obtained during the training stage. Thus, the underlying distribution of the error could be used to compute specific confidence intervals with the desired signification for each sensor. On the one hand, the absolute mean error of the sensors measurements properly fits a normal distribution. On the other hand, the mean squared error fits an exponential distribution. 

Therefore, the confidence intervals were generated by using the normal distribution associated to the error of each sensor. This choice was based on the fact that the absolute mean error was approximately equal to the uncalibration that was taking place. 

Finally, a rejection takes place whenever the value exceeds the bounds of the confidence interval. The density of rejections is computed as the rolling ratio between the number of rejections for a window of a fixed time length (with 1440 minutes as default value, which corresponds to the number of minutes in a day) and the length of the window. 

The rejection density is the variable that triggered the uncalibration warning. A uncalibration is said to take place whenever a threshold is overcome during a specified amount of time.

It should be considered that the proposed architecture does not really know what being calibrated means. The expected behaviour is that the \ac{NN} detects small differences between the current measurement and the one from which it learnt. Thus, in the eventuality of a maintenance task in any sensor, the architecture proposed might detect this new event as a potential uncalibration. This uncalibration event will remain active until this new condition of the sensor is re-learnt as ‘calibrated’ by the model.

The standard procedure to overcome this issue is to re-train the model under these new conditions. Thus, this process would require a very high temporal cost, since collecting data during almost a whole year would be necessary. Because of this reason, transfer learning arises as a feasible solution. Using transfer learning, a small amount of data is needed to teach the new condition to the \ac{NN}.

\subsection{Transfer learning}

Transfer learning \cite{bozinovski2020reminder} is a \ac{ML} technique aimed at specializing \ac{ML} models with a minimum amount of data. This technique consists of two main steps. The first step is training a model for a more general task conceptually related with the target or specialized task. For instance, the target task could be identifying a specific face in a picture and, in that case, the general task would be training a model for general face identification. The second step, would be to re-train a subset of layers of the model with a training dataset made out of elements corresponding to the specialized task (a specific face in the example presented).

In the case of this work, multiple scenarios can benefit from transfer learning. These scenarios can be summarized as it follows: changing the ground truth value for all sensors or for a subset of them, including new sensors and adapting to some other environments with, possibly, lack of information.

As it was presented in subsection \ref{subsec:dataset}, a different slicing of the dataset was used for transfer learning. Transfer learning is applied by dividing the original dataset into three smaller subsets, namely, a training subset for the original model, which will be called model A, a re-training subset for the application of transfer learning (model A will be renamed to model B after this process) and, finally, a testing subset for both models. Regarding the training subset, it contains 70\% of the datapoints. This stage corresponds to the training for the general task. Regarding the subset devoted to be used for re-training the model and, thus, for the application of transfer learning, it contains the 20\% of the datapoints. The amount of required datapoints can be reduced to up to 2.5\% of the total (10000 datapoints per sensor) in the case of temperature and humidity. Finally, the testing subset includes 10\% of the datapoints. This subset was used to test the performance of the neural network, both before and after the application of the transfer learning. A constant offset was added in the last part of the test set in order to simulate the recalibration due to a maintenance task. It is important to note that this offset is not applied to the whole test set in order to properly see the change of conditions and the associated evaluation by both, model A and model B.

As mentioned, one of the main purposes of the application of transfer learning is the reduced amount of necessary samples for training a model. This amount is directly associated to the number of parameters to retrain, which, at the same time, is intimately related to the number of layers to retrain. The nature of the target task, i.e., how abstract it is, and where (in which layers) the concepts learnt are located within the \ac{NN}, are crucial factors which will determine the amount of required data. During the current research, transfer learning has been applied on the very last hidden layer of the \ac{NN}, i.e., only the weighs connecting the last hidden layer with the output layer has been retrained. The obtained results suggest that training this subset of weights was enough for the existing requirements.

\section{Results}
\label{sec:results}
This section shows the results obtained using the methods presented in section \ref{sec:methods}. Two different experiments were conducted: (1) to evaluate the capability for the detection of uncalibrations of the current approach through, which has been called, performance evaluation experiments; (2) to test the flexibility and scalability of the solution, which have been called transfer learning experiments.

\subsection{Performance evaluation experiments}
\label{subsec:mlexper}
These experiments were conducted to test the detection of uncalibrated sensors using architecture presented. The dataset used is the one shown in \ref{subsec:dataset}. Uncalibrations were introduced in the testing set as drifts of different nature (e.g linear, exponential or logarithmic) across time and in different temporal frames (for instance, at the beginning, in the middle and at the end of the year).

\begin{figure}[h!]
\centering
\includegraphics[width=0.9\textwidth,keepaspectratio]{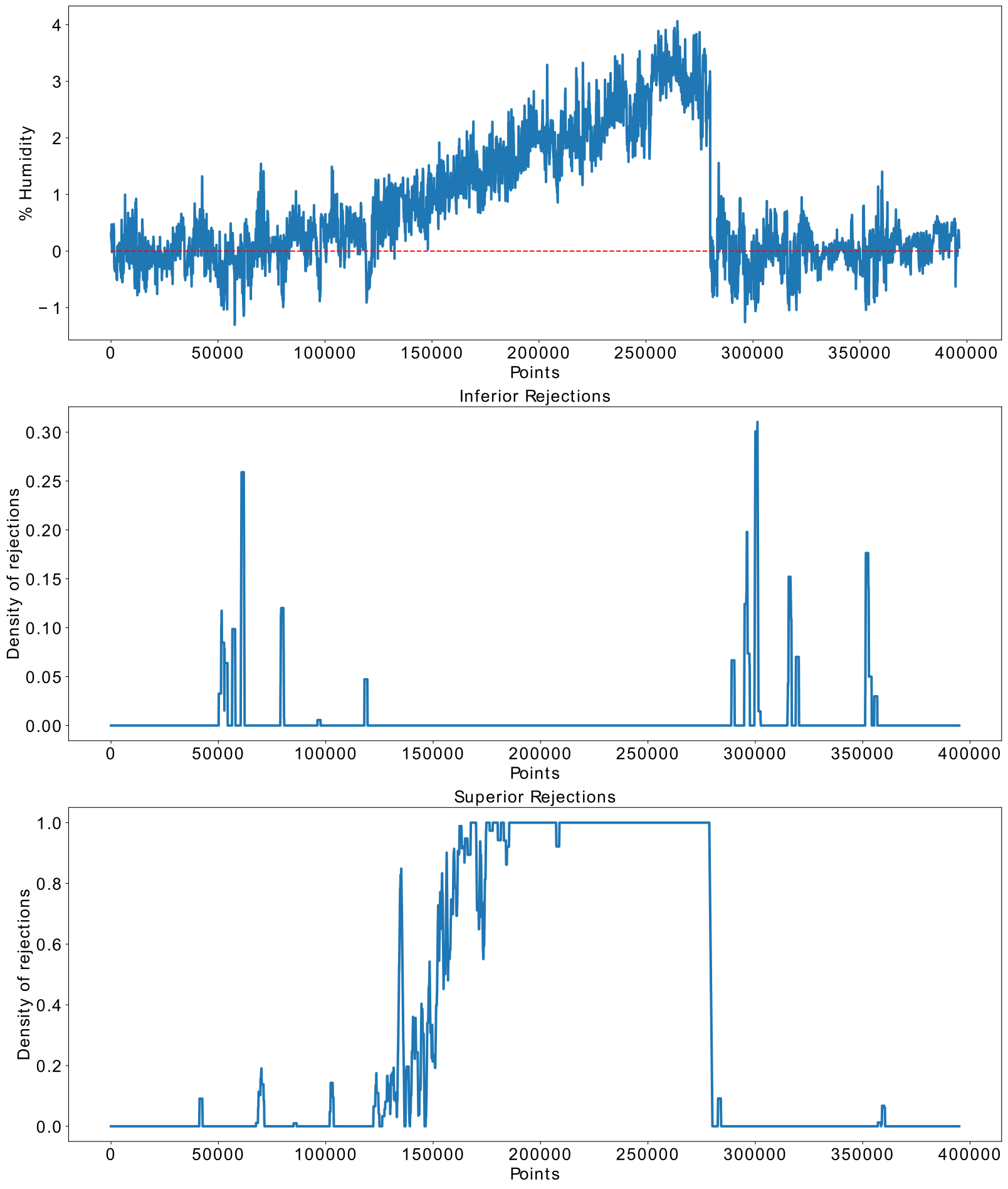}
\caption{Uncalibration detection through rejection density. In the upper subplot, the blue trace shows the error corresponding to a linear uncalibration and the red dashed line stands for the error zero value. In the middle subplot, the value of the rejection density for error values exceeding the lower bounds of the confidence interval is shown. As it can be seen, the maximum value for this rejection is, approximately, 0.3. In the lower subplot, the value of the rejection density for error values exceeding the upper bounds of the confidence interval is shown. As it can be seen, this density reaches the maximum possible value from a critical point and during the whole uncalibration phenomenon.}
\label{Rejection Density}
\end{figure}

An uncalibration is considered to be detected once the rejection density value has overcome a predefined threshold, which depends on the sensor, and whose average value is 0.8. Besides, this threshold has to be overcome during a predefined period of time, which also depends on the sensor, and can be defined as two weeks, since it is enough time for all sensors and it is a reasonably low period of time for uncalibration detection. An example of the application of this technique can be seen in figure \ref{Rejection Density} where rejection density for both the upper and lower bound of the confidence interval are shown. 

The second and third plots show the rejection density values. It can be seen that these values keep low whenever a uncalibration is not taking place. Conversely, when a uncalibration takes place, the rejection density reaches values close to one.  

Regarding the performance of the approach, uncalibrations were detected for all different types of sensors. However, two different scenarios can be distinguished, namely, a first optimistic scenario, where the uncalibration is detected within the tolerance ranges defined by the quality requirements of the pharma industry (i.e. $\pm 0.5$\textdegree,$\pm 3\%$ and $\pm 0.5$P in temperature, humidity and pressure respectively), and a non-optimistic scenario, where the uncalibration is detected over the tolerance range. For all those sensors that are in the optimistic scenario, an estimation of the remaining time until the tolerance range gets exceeded can be provided. 

Thus, for the case of the temperature sensors, the proposed approach was tested on the available data coming from temperature sensors over 17 rooms. All uncalibrations were detected for all sensors. In the best scenario, the architecture detected uncalibrations associated to deviations of 0.25\textdegree C (tolerance 0.5\textdegree C). This accuracy was reached for 16 rooms out of 17 of the second floor. In the worst scenario, the uncalibration was detected for deviations of, approximately, 0.5\textdegree C. 

For the humidity sensors case, the proposed method was tested on the available data coming from humidity sensors over 17 clean rooms. All uncalibrations were detected for all sensors. In the best scenario, the algorithm detects uncalibrations associated to deviations of 2\% (tolerance 3\%). This accuracy was reached for 14 rooms out of 17 of the second floor. In the worst scenario, the uncalibration was detected for deviations close to 3\%. 

Finally, for the pressure sensors case, the proposed method was tested on the available data coming from over 24 pressure sensors. All uncalibrations were detected for all sensors. In the best scenario, the algorithm detected uncalibrations associated to deviations under 0.5 Pascals (tolerance 0.5 Pascals). This accuracy was reached for 6 out of 20 rooms of the second floor. In the worst scenario, the uncalibration was detected for deviations close to 1.5 Pascals.

\begin{figure}[h!]
\centering
\includegraphics[width=\textwidth,keepaspectratio]{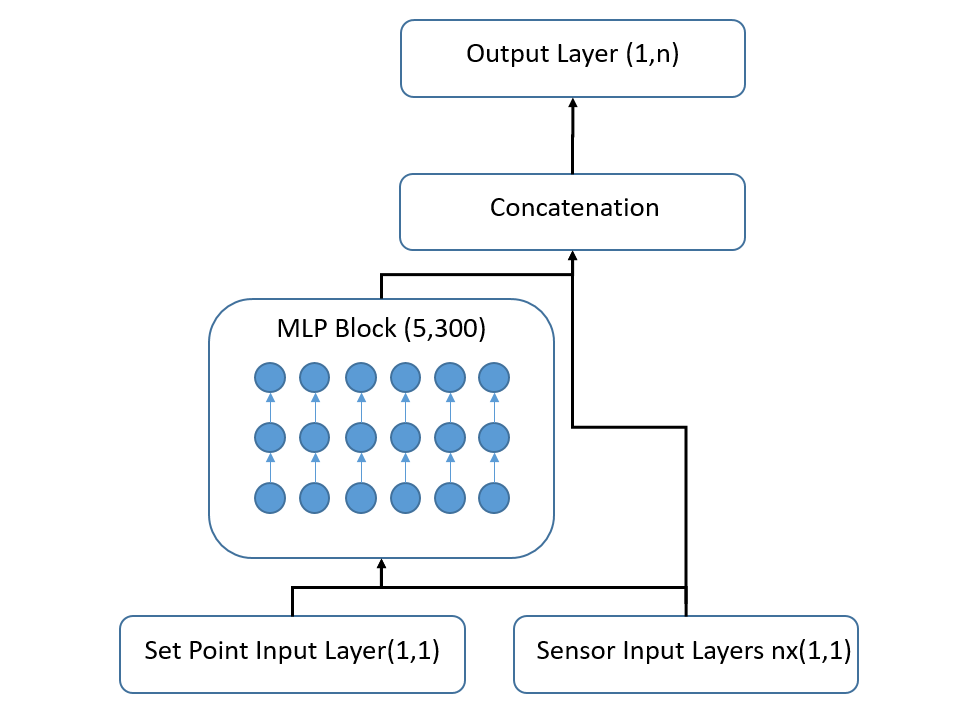}
\caption{Wide and Deep model diagram. In this figure, the tested architecture of the wide and deep model is shown. It can be seen $n+1$ different input layers; an input layer with one single neuron, which is connected to a hidden block of five hidden layers with three hundred neurons each. This input layer takes the estimated global set point as input. Next, $n$ input layers with one single neuron each. Each input layer takes the measurements coming from the different associated sensors as input. Finally, these input layers are concatenated to the output of the hidden block and connected to the output layer, which has a neuron per sensor.}
\label{WDARchitecture}
\end{figure}

Some other \ac{ANN} architectures were tested during the development of the present research. More precisely, the same experiments were conducted using a Wide \& Deep Neural Network for temperature and humidity sensors and \ac{RNN} for the pressure sensors. The architecture of the Wide \& Deep Neural Network was based on the \ac{MLP} used on the implemented approach, but also includes an input layer per sensor, so that the precise information coming separately from each sensor can be included. This can be seen in figure \ref{WDARchitecture}.

Wide and Deep architecture yielded better results for most of the rooms. However, for two of them, the uncalibrations were not detected. It has been observed that, for these two rooms, how the sensor is supposed to behave under a specific condition is not learnt by the neural network, and the identity function is approximated instead. Thus, when an anomaly of any kind was introduced on the input signal, this same value was retrieved by the \ac{NN} and, therefore, the uncalibration was not detected. It has been observed that the worst results were obtained for these two rooms independently of the architecture. This fact suggests that the local conditions inside these rooms had a deep impact on the learning capabilities of the different models. 

Regarding the \ac{RNN}, \ac{LSTM} neurons were used. In this case, uncalibrations were not detected either. This was probably caused due to the long-term nature of uncalibrations. Uncalibrations are long term phenomenons, thus, the deviation associated to an uncalibration will be observed during long periods of time. Hence, the deviation will be included as relevant by \ac{LSTM} neurons during the information inclusion stage, that is, when the input gate evaluates what information is relevant and what information is not. In this case, the learning of this long-term phenomenons by the \ac{NN} provoke the worst results. 

\subsection{Transfer learning experiments}
\label{subsec:tlexper}
These experiments were conducted to test the capability of the architecture for learning a new condition with the less temporal cost. Thus, these experiments were split in three categories, namely, change in global or specific conditions, adding new sensors and adapting to some other environments with lack of information. To evaluate transfer learning, the error obtained for the main model (model A) and the retrained model (model B) under the specific new condition were compared.

\subsubsection{Change of conditions}
These experiments were conducted to test the capability of the architecture proposed for learning a new condition as the current calibration status for one or more sensors. 

\begin{figure}[h!]
\centering
\includegraphics[width=\textwidth,keepaspectratio]{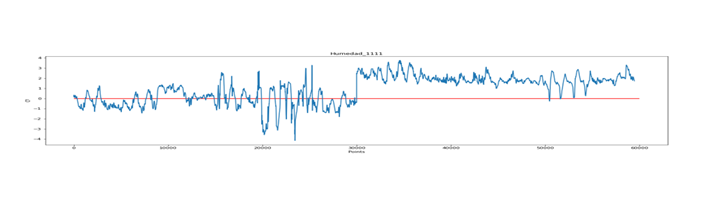}
\caption{In this figure, the error associated to a scheduled recalibration can be seen. The blue trace shows the error and the red dashed line stands for the error zero value. The recalibration takes place at the mid point of the figure. Since transfer learning is not applied, once the recalibration takes place, the error immediately increases.}
\label{individualtransnotrain}
\end{figure}

In the case of the individual offset, in figure \ref{individualtransnotrain} two different phases of the error obtained from model A can be observed. The first one corresponds to the model evaluation under normal conditions. The second one corresponds to the model evaluation under the generation of a specific offset. It can be seen how the error is displaced, approximately, three units which is the same amount of the generated offset. 

\begin{figure}[h!]
\centering
\includegraphics[width=\textwidth,keepaspectratio]{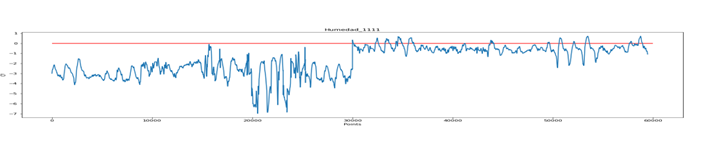}
\caption{In this figure, the blue trace shows the error associated to a scheduled recalibration and the red dashed line stands for the error zero value. In this case, transfer learning is applied, thus, once the recalibration takes place, the error immediately decreases. It is important to note that this model is not applied in the previous moments to the recalibration task. }
\label{individualtranstrain}
\end{figure}

However, regarding model B, it can be seen how the error returns to the same range of values as previously. This behaviour suggests that the expected results have been obtained. 
This experiment was reproduced for different rooms and the different types of sensors, obtaining similar results. It can be observed in figure \ref{individualtranstrain}.

It should be noted that the minimum number of datapoints to properly use transfer learning is around ten thousand samples. This number of samples corresponds, approximately, to one week of data collection.


Regarding the generation of an offset for the whole set of sensors, the model did not detect any change in its behavior. It suggests that, if the behaviour of the whole set of sensors is equally changed, then, no effect on the joint behaviour can be detected. Thus, no uncalibration is taking place from the point of view of the model. This fact proposes an interesting question regarding how the joint behaviour could be learnt.

\subsubsection{Adding new sensors to the model}
These experiments were conducted to determine the time that the architecture takes to learn and include the behaviour of new sensors in the loop. These tests were made using the humidity and temperature sensors.

\begin{figure}[H]
\centering
\includegraphics[width=\textwidth, height=0.7\textheight,keepaspectratio]{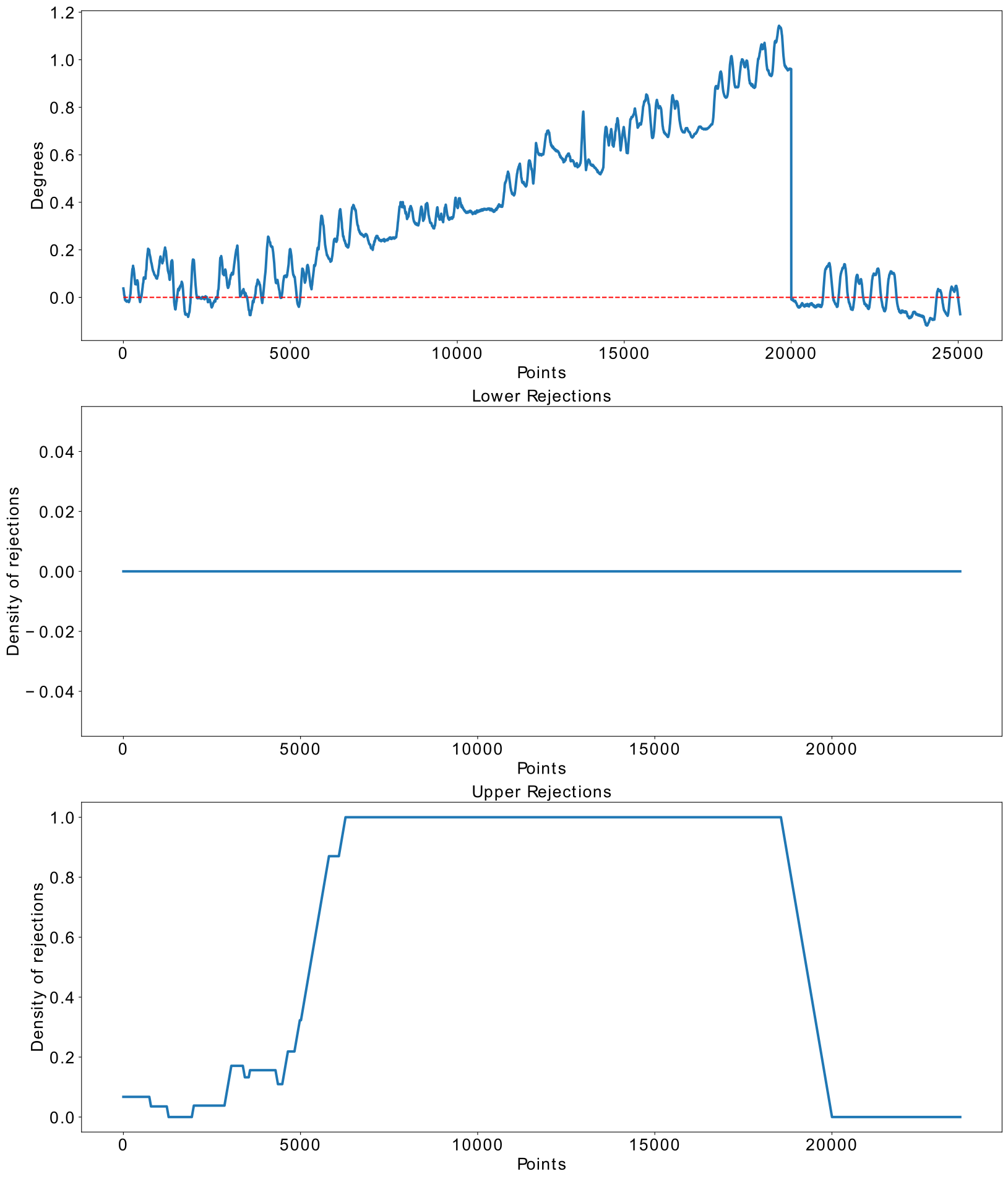}
\caption{Uncalibration detection for new included sensors. In this figure, we can see the results of the application of transfer learning in order to include new sensors in the solution. The detection of an uncalibration for one of those sensors is shown. In the upper subplot,the error corresponding to a linear uncalibration can be seen. The red dashed line stands for the error zero value. In the middle subplot, the value of the rejection density for error values exceeding the lower bounds of the confidence interval is presented. In this case, the maximum value for this rejection is approximately zero. In the lower subplot,  the value of the rejection density for error values exceeding the upper bounds of the confidence interval is presented. As it can be seen, this density reaches the maximum possible value from a critical point and during the whole uncalibration phenomenon.}
\label{ModelB3}
\end{figure}

A first dataset with data coming from 13 sensors out of seventeen was taken as main set. For this set of sensors, $o_1 = 322677$ datapoints were available. In this case, model A was trained with this dataset. Next, model B was re-trained with data coming from both the initial set of thirteen sensors and a set of four new sensors. The obtained results show that uncalibrations on the new set of sensors can be properly detected by model B. Figure \ref{ModelB3} shows how uncalibrations are detected in the new sensors that were included for the temperature case. 

\subsubsection{Adapting to some other environment with lack of information}
These experiments were conducted to test if the architecture could be trained with data coming from one specific location and then be implemented on a different location after a transfer learning process.

\begin{figure}[H]
\centering
\includegraphics[width=\textwidth, height=0.7\textheight,keepaspectratio]{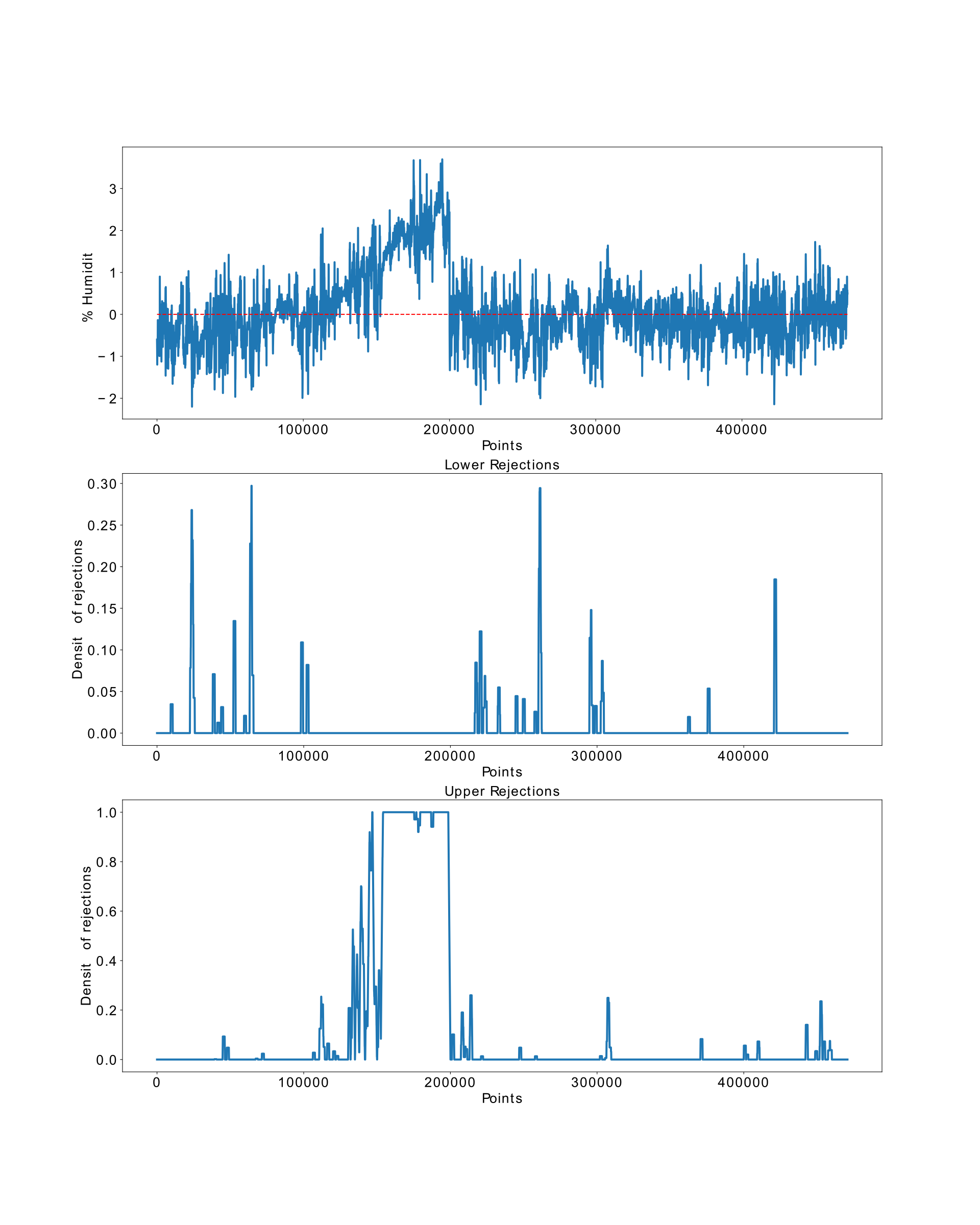}
\caption{Uncalibration detection in the basement. In this figure, the detection of an uncalibration taking place in the basement is shown. In this case, transfer learning has been applied in order to re-train a model with a small quantity of data coming from the basement. The model was previously trained with data coming from the second floor. In the upper subplot, the error corresponding to a linear uncalibration is shown in blue. The red dashed line stands for the error zero value. In the middle subplot, the value of the rejection density for error values exceeding the lower bounds of the confidence interval is shown. In this case, the maximum value for this rejection is 0.3. In the lower subplot, the value of the rejection density for error values exceeding the upper bounds of the confidence interval is shown. As it can be seen, this density reaches the maximum value starting from a critical point and remains during the whole uncalibration phenomenon.}
\label{ModelOG02OG01}
\end{figure}

In this case, model A was trained with data coming from the second floor with 17 sensors of both types, temperature and humidity. Then, model B was re-trained with a small amount of data coming from the basement. Approximately, 80000 datapoints were used. 
The obtained results show that uncalibrations are indeed detected by model B in this new environment. Furthermore, the problem of the lack of data was solved by applying this method. Figure \ref{ModelOG02OG01} shows an example of how uncalibrations were detected by model B in this specific context.

The obtained results suggest that transfer learning is a valid technique for the adaptation of the solution to some other locations. 

\section{Discussion and Conclusions}
\label{sec:discussion}

Regarding the performance of the solution, it can be firstly concluded that uncalibrations can indeed be detected even for values smaller than the defined tolerance ranges for most cases in temperature and humidity sensors.

Secondly, it can be concluded that the application of transfer learning yielded results that were coherent with the expected ones. Besides, these results empowered the flexibility and scalability of the solution, enabling the use of the model in a wide variety of contexts such as sensor addition, integration within new environments and partially solving problems associated to the lack of data.


Furthermore, transfer learning retrieved good results for very low amounts of data. The minimum amount of  data required is about 10000 samples which is the data coming from one week of observations at a sampling rate of one sample per minute. Indeed, the fact that only the last weight matrix was needed to re-train means that the information associated to all these modifications were learned on a very abstract level in the \ac{NN} \cite{geron2019hands}. This fact suggest, on the one hand, that offsets have a low impact on the joint dynamics of sensors. Thus, the joint behavior of the whole set of sensors is ruled by much deeper relationships than the addition of specific values. On the other hand, the results obtained for pressure sensors, despite they are not under a common constant condition, may be explained by this fact. The sudden changes in pressure values could be then understood as an offset, what implies that the model has a deep knowledge of the sensor behavior, although these offsets reduce the accuracy on the detection of uncalibrations.

As a remarkable aspect of this research, it is important to note that the solution has been trained and tested in a real context, with data coming directly from production. Indeed, the solution presented in this work is currently deployed in Azure \cite{chappell2010azure}, where the solution performs the uncalibration detection directly from the information of the sensors from a digital twin of the building. This digital twin contains the current status of the physical sensors.



To the best of our knowledge, this is the first time that potential uncalibrations of a set of sensors are online detected whenever the set point is unknown. Furthermore, the proposed architecture can be extended by means of transfer learning to a wide range of different fields.

Regarding the future work, different targets are considered: (1) extending the results obtained after the application of transfer learning. Thus, we are interested in training the solution with a generic set of sensors of one type (such as temperature sensors) and then, testing the performance of the solution on sensor systems of a different kind, for instance, engine sensor systems, smart-city sensors systems and power plants sensor systems, among others; (2) obtaining information about the learning of the solution through the application of \ac{XAI}. This could provide useful information about the potential scalability and flexibility of the solution and the different models to which it could be applied.  

\section*{Acknowledgement}
This work was partially supported by the Altran Innovation Center for Advance Manufacturing, the EU H2020 project Virtual IoT Maintenance System (VIMS Grant ID: 878757), the Spanish grant (with support from the European Regional Development Fund) MIND-ROB (PID2019-105556GB-C33) and by the EU H2020 project CHIST-ERA SMALL (PCI2019-111841-2). 

\bibliographystyle{elsarticle-num-names}
\bibliography{references.bib}

\end{document}